\documentclass[letterpaper, 10 pt, conference]{ieeeconf}  % Comment this line out if you need a4paper

\IEEEoverridecommandlockouts                              % This command is only needed if 
\usepackage{booktabs}
\usepackage{siunitx}

\usepackage{multirow}
\usepackage[normalem]{ulem}
\usepackage{hyperref}
\usepackage{amssymb}
\usepackage{bm}
\usepackage{bbm}

\usepackage{amsmath}
\usepackage{graphicx}
\usepackage{url}
\usepackage{algorithm2e}

\usepackage[T1]{fontenc}
\usepackage{dsfont}

\title{\LARGE \bf
SeqRisk: Transformer-augmented latent variable model for robust survival prediction with longitudinal data
}

\author{Mine Öğretir$^{1,2}$, Miika Koskinen$^{2,3}$, Juha Sinisalo$^{2,3}$, Risto Renkonen$^{3}$ and Harri Lähdesmäki$^{1}$% <-this % stops a space
\thanks{$^{1}$Computer Science Department, Aalto University, Espoo, Finland}
\thanks{$^{2}$ HUS Helsinki University Hospital, Helsinki, Finland}
\thanks{$^{3}$ University of Helsinki, Helsinki, Finland}
}

\begin{document}

\maketitle
\thispagestyle{empty}
\pagestyle{empty}

\begin{abstract}
In healthcare, risk assessment of patient outcomes has been based on survival analysis for a long time, i.e.\ modeling time-to-event associations. However, conventional approaches rely on data from a single time-point, making them suboptimal for fully leveraging longitudinal patient history and capturing temporal regularities. Focusing on clinical real-world data and acknowledging its challenges, we utilize latent variable models to effectively handle irregular, noisy, and sparsely observed longitudinal data. We propose SeqRisk, a method that combines variational autoencoder (VAE) or longitudinal VAE (LVAE) with a transformer-based sequence aggregation and Cox proportional hazards module for risk prediction. SeqRisk captures long-range interactions, enhances predictive accuracy and generalizability, as well as provides partial explainability for sample population characteristics in attempts to identify high-risk patients. SeqRisk demonstrated robust performance under conditions of increasing sparsity, consistently surpassing existing approaches.

\end{abstract}

% \begin{keywords}
% survival analysis, time-to-event data, longitudinal measurements, deep learning, VAE, transformer, EHR
% \end{keywords}

\section{Introduction}
\label{sec:intro}

Survival analysis comprises methods for time-to-event outcomes and plays a central role in healthcare for prognosis, monitoring disease progression, evaluating treatments, and identifying high-risk subgroups \cite{ambale2017cardiovascular,ghosh2021efficient}. Modern clinical datasets pose challenges---high dimensionality, irregular sampling, and substantial missingness---that strain classical models such as the Cox proportional hazards model \cite{cox1972regression} and traditional machine learning approaches like random survival forests \cite{ishwaran2008random}.

Deep learning has advanced survival modeling beyond linear effects, first in static settings with models such as \emph{DeepSurv} \cite{katzman2018deepsurv} and \emph{DeepHit} \cite{lee2018deephit}, and more recently through approaches that leverage variational autoencoders (VAEs), transformers, or neural ordinary differential equations (ODEs) \cite{kim2020improved,manduchi2021deep,mesinovic2024dysurv,hu2021transformer,wang2022survtrace}. However, many existing methods either treat observations as independent or require regular sampling, limiting their applicability to real-world longitudinal electronic health records (EHRs), where measurements are often irregular, incomplete, and high-dimensional.

In this work, we propose \emph{SeqRisk}, a unified generative--discriminative framework for longitudinal survival prediction. Our method learns latent trajectories from irregularly sampled multivariate data using a variational autoencoder (VAE), or a longitudinal VAE (LVAE) that is augmented with an additive multi-output Gaussian process prior \cite{kingma2013auto,ramchandran2021longitudinal}. A transformer-based module \cite{vaswani2017attention} aggregates these latent sequences, while a proportional hazards head yields risk scores using a nonlinear extension of the Cox model \cite{katzman2018deepsurv}. Crucially, SeqRisk is trained end-to-end with a joint objective combining the evidence lower bound (ELBO) and Cox partial likelihood, encouraging latent representations that are both structured for irregular longitudinal data and aligned with hazard prediction.

\textbf{Contributions.} While VAE-based survival representation learning and transformer-based survival models have been explored separately, SeqRisk integrates these directions into a single generative--discriminative framework for irregular longitudinal EHRs. Specifically, we contribute:
(i) a unified latent layer, instantiated as either a standard VAE (for i.i.d.\ representations) or an LVAE (for irregular longitudinal data), providing flexible latent encodings of multivariate patient trajectories;
(ii) a transformer-based survival head that captures long-range temporal dependencies for time-to-event prediction;
(iii) an end-to-end training objective that combines generative modeling (via the ELBO) with supervised survival learning (via the Cox partial likelihood), yielding latent representations that are both reconstruction-faithful and hazard-discriminative.

\section{Related Work}
\label{sec:related}

% \paragraph{Static deep survival.}
Many neural survival models focus on static covariates. \emph{DeepSurv} models nonlinear proportional hazards with a fully connected network \cite{katzman2018deepsurv}, while \emph{DeepHit} predicts discrete-time survival probabilities via a multitask loss \cite{lee2018deephit}. Extensions to these approaches have incorporated variational or transformer-based representations for static covariates \cite{kim2020improved},\cite{manduchi2021deep},\cite{mesinovic2024dysurv},\cite{hu2021transformer},\cite{wang2022survtrace},\cite{nagpal2021deep}, but they typically require pre-aggregated data and fail to fully exploit the temporal structure in longitudinal records.

% \paragraph{Longitudinal and recurrent deep models.}
\emph{Recurrent Deep Survival Machines} combine RNNs with parametric survival heads \cite{nagpal2021recurrent}, but rely on strong distributional assumptions, e.g.\ Weibull or Log-normal, and struggle with irregular sampling. \emph{DeepJoint} jointly models visits, missingness, and survival events with a multi-task RNN, yet it imposes complex training requirements and largely deterministic treatments of missingness \cite{jeanselme2022deepjoint}. More recently, models based on neural ODEs have introduced continuous-time dynamics. For instance, \emph{SurvLatent ODE} \cite{moon2022survlatent} encodes EHR sequences with ODE-RNNs and models competing risks, while \emph{SODEN} \cite{tang2022soden} treats survival as an ODE-constrained optimization problem. Neural controlled differential equations (CDEs) \cite{kidger2020neural} further enhance the modeling of asynchronous measurements. \cite{bleistein2024dynamical} present a survival model based on neural CDEs and signature kernels that captures irregular sampling patterns while providing theoretical guarantees, though it introduces additional computational overhead.

% \paragraph{Transformer-based joint models.}
Recent approaches incorporate transformers into joint modeling frameworks for longitudinal survival prediction. \emph{TransformerJM} \cite{lin2022deep} combines transformer layers with time-aware encoding for survival, although it typically assumes regular sampling. \emph{TransformerLSR} \cite{zhang2025transformerlsr} jointly learns longitudinal and survival processes via an autoregressive trajectory representation and deep point process, showing strong performance but with complex modeling assumptions. Other methods such as \emph{STRAFE} \cite{zisser2024strafe} and \emph{VaDeSC-EHR} \cite{qiu2025vadesc} address scalable survival modeling or latent clustering, but often rely on rigid data assumptions or simplified survival distributions. \emph{UniSurv} \cite{zhang2025unisurv} focuses on continuous-time risks but introduces unimodality constraints or high computational complexity.

% \paragraph{Positioning of this work.}

Our proposed method differs from the approaches above in two key aspects: (i) it can learn patient-specific latent trajectories using a VAE with a Gaussian process prior, explicitly modeling temporal correlations from irregular observations; (ii) it uses a transformer-based Cox proportional hazards head to predict time-to-event outcomes from these latent sequences. This design couples generative representation learning with survival prediction while enabling end-to-end training. Unlike transformer-only survival models, our method retains a rich generative latent space regularized by reconstruction loss. To our knowledge, this is the first method to integrate LVAE-based latent modeling with transformer survival prediction under the Cox framework in an end-to-end differentiable setup. This combination provides a flexible, data-efficient, robust, and principled solution for longitudinal survival analysis on irregular and high dimensional clinical data. 

% geri eklenecek
\begin{figure}
  \centering
  \includegraphics[width=.7\linewidth]{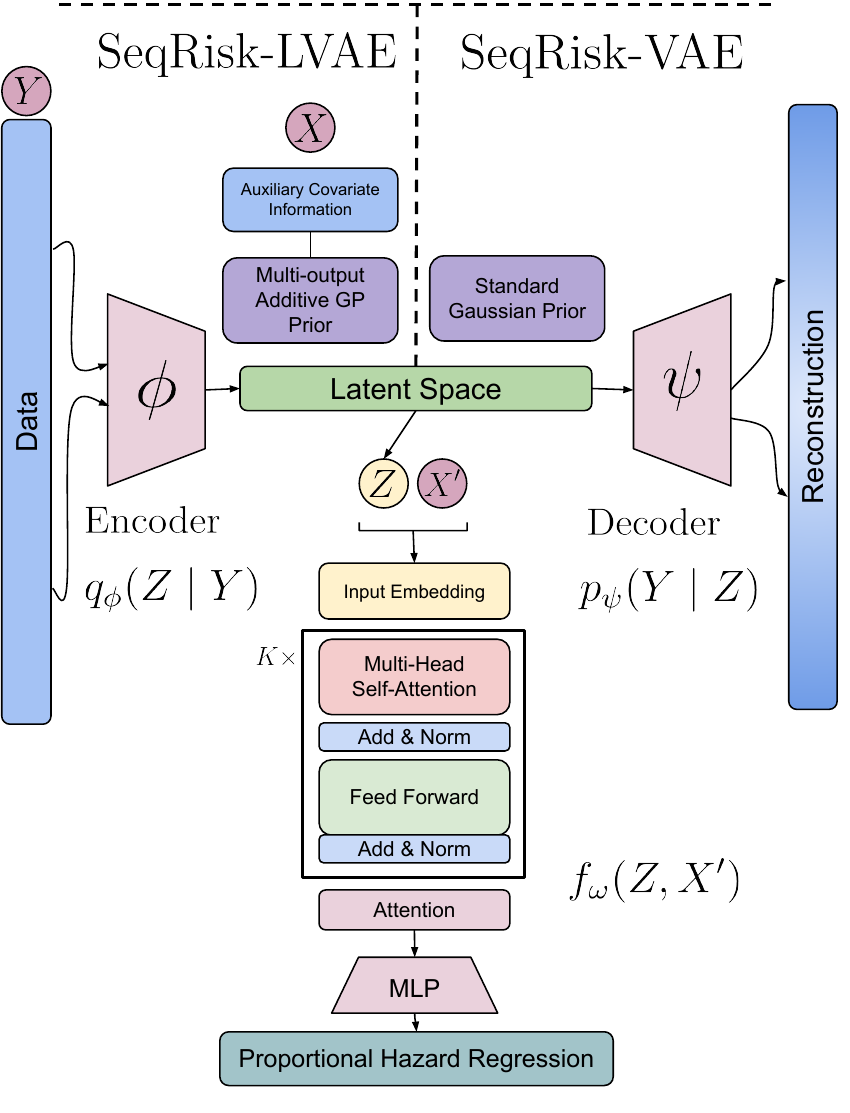}
  \caption{Overview of SeqRisk}
  \label{fig:model}
\end{figure}

\section{SeqRisk Model}
\label{sec:seqrisk_framework}

The SeqRisk model is designed to enhance the predictive capabilities of time-to-event analysis using advanced machine learning techniques. This model integrates a VAE --- either standard VAE or longitudinal VAE (LVAE) --- with a transformer-based sequence aggregation module, leading to a proportional hazard regression to estimate survival risks (see Figure \ref{fig:model} for an overview). %

\paragraph{Notation} 
Let \(P\) represent the total number of distinct instances (such as individuals), each with \(n_p\) time-series samples. The number of all longitudinal samples across instances is denoted by \(N = \sum_{p=1}^{P} n_{p}\). For each individual \(p\), we have data $(X_p, Y_p, t_p, e_p)$, where \(X_{p} = [\boldsymbol{x}_{1}^{p}, \ldots, \boldsymbol{x}_{n_{p}}^{p}]\) denotes covariate data, including e.g.\ the measurement times and patient demographics,  \(Y_{p} = [\boldsymbol{y}_{1}^{p}, \ldots, \boldsymbol{y}_{n_{p}}^{p}]\) denotes measurement variables, e.g.\ lab results, \( t_p \) is the time-to-the-event or censoring, and \( e_p \) is the event indicator with value 1 for event and 0 for censoring. The collective longitudinal data across all instances is represented as \( \{ (X_p, Y_p, t_p, e_p) \}_{p=1}^P \).

The domain of covariates \(\boldsymbol{x}_i^p\) is defined by \(\mathcal{X} = \mathcal{X}_{1} \times \ldots \times \mathcal{X}_{Q}\), where \(Q\) indicates the total number of covariates, and \(\mathcal{X}_{q}\) corresponds to the domain of the \(q\)\textsuperscript{th} covariate, which may be continuous, categorical, or binary. The domain of \(\boldsymbol{y}_i^p\) is defined by $\mathcal{Y}=\mathbb{R}^D$, where $D$ denotes the number of measurement variables per observation.
% Furthermore, the latent embedding of the $N$ samples \(Y = [Y_1,\ldots,Y_P] = [\boldsymbol{y}_1,\ldots,\boldsymbol{y}_N]\) are in an \(L\)-dimensional vector space represented by \(Z = [Z_1,\ldots,Z_P] = [\boldsymbol{z}_{1}, \ldots, \boldsymbol{z}_{N}] \in \mathbb{R}^{N \times L}\). %
Furthermore, we denote the $L$-dimensional latent space by $\mathcal{Z}=\mathbb{R}^L$ and the latent embedding for all $N$ samples by
$Z=[\boldsymbol{z}_1,\ldots,\boldsymbol{z}_N]^\top\in\mathbb{R}^{N\times L}$, where $\boldsymbol{z}_i\in\mathbb{R}^L$ is the latent vector for sample $i$.
We also index $Z$ by latent dimensions as $Z=[\tilde{\boldsymbol{z}}_1,\ldots,\tilde{\boldsymbol{z}}_L]$, where
$\tilde{\boldsymbol{z}}_l=[z_{1l},\ldots,z_{Nl}]^\top\in\mathbb{R}^N$ collects the $l$-th latent dimension across all $N$ samples.
Covariates across all $N$ samples are denoted similarly as $X = [X_1,\ldots,X_P]$. We flatten all observations across individuals into $\{(\boldsymbol{x}_i,\boldsymbol{y}_i)\}_{i=1}^N$ by concatenating visits over $p$ (so index $i$ corresponds to the $j$-th sample of some individual $p$).

\subsection{Model Architecture}
\label{subsec:model_architecture}

The SeqRisk model combines a latent generative module (VAE/LVAE) with a transformer-based survival head.

% The SeqRisk model employs a dual-model approach to address the diverse needs of survival analysis:

\paragraph{Variational Autoencoder (VAE)}
We use a standard VAE \cite{kingma2013auto} to learn low-dimensional latent representations $Z$ from high-dimensional measurements $Y$.
% As a generative model, VAE \cite{kingma2013auto} is adept at learning complex distributions from high-dimensional data. We use the standard VAE to learn the variational approximations of latent variables without temporal considerations. 
% The standard VAE serves as the foundation for learning latent representations $Z$ from high-dimensional input data $Y$. 
Training VAEs involves minimizing the Kullback-Leibler (KL) divergence of amortized variational
approximation, 
% \( q_{\phi}(\boldsymbol{z} |  \boldsymbol{y}) \), 
from the posterior of the latent variable $\boldsymbol{z}$ given the observed sample $\boldsymbol{y}$, $p(\boldsymbol{z} |  \boldsymbol{y})$, that corresponds to maximizing the evidence lower bound (ELBO). For the full dataset $Y$ the ELBO is %
\begin{align}
\label{ELBO_VAE}
\log p_{\psi}(Y) &\geq \mathcal{L}_{\phi, \psi}(Y) \\
&= \sum_{i=1}^N \mathbb{E}_{q_{\phi}(\boldsymbol{z}_i|\boldsymbol{y}_i)}[\log p_{\psi}(\boldsymbol{y}_i|\boldsymbol{z}_i)] \nonumber \\
& \quad - D_{\mathrm{KL}}(q_{\phi}(\boldsymbol{z}_i|\boldsymbol{y}_i) \parallel p(\boldsymbol{z}_i)). \nonumber
\end{align}
In Eq.~(\eqref{ELBO_VAE}), $q_{\phi}(\boldsymbol{z}\mid\boldsymbol{y})$ is the encoder-defined variational posterior and $p_{\psi}(\boldsymbol{y}\mid\boldsymbol{z})$ is the decoder likelihood, with parameters $\phi$ and $\psi$; we use a standard normal prior $p(\boldsymbol{z})=\mathcal{N}(\mathbf{0},\mathbf{I})$.
% This objective includes the expected log likelihood, enhancing data reconstruction fidelity, and the KL divergence, which serves as a regularizer by maintaining the distributional integrity of the latent space.

\paragraph{Longitudinal Variational Autoencoder (LVAE)}

While VAE assumes i.i.d.\ samples, LVAE allows the model to account for patient-specific variations and temporal dynamics,
making it particularly suited for longitudinal data analysis \cite{ramchandran2021longitudinal}. 
LVAE places an additive Gaussian process (GP) prior over each latent-dimension function $f_l(\cdot)$ conditioned on covariates $\boldsymbol{x}$:
$f_l(\cdot)\sim \mathcal{GP}\!\big(0,\,k_l(\cdot,\cdot\mid\theta)\big), \qquad l\in\{1,\dots,L\},$
where $k_l(\cdot,\cdot\mid\theta)$ is a positive semi-definite kernel with hyperparameters $\theta$ (e.g., lengthscales and variance).
We use an additive kernel over $R$ covariate subsets $\{\boldsymbol{x}^{(r)}\}_{r=1}^R$ (e.g., \{$subject$ $index$\}, \{$age$\} and \{$gender$, $age$\}),
\[
k_l(\boldsymbol{x},\boldsymbol{x}')=\sum_{r=1}^R k_l^{(r)}(\boldsymbol{x}^{(r)},\boldsymbol{x}'^{(r)}),
\]
equivalently $f_l(\boldsymbol{x})=\sum_{r=1}^R f_l^{(r)}(\boldsymbol{x}^{(r)})$ with $f_l^{(r)}\sim\mathcal{GP}(0,k_l^{(r)})$.
Over the $N$ samples with covariates $X=\{\boldsymbol{x}_i\}_{i=1}^N$, this yields
$\tilde{\boldsymbol{z}}_l=[f_l(\boldsymbol{x}_1),\ldots,f_l(\boldsymbol{x}_N)]^\top \in\mathbb{R}^N$ and
$\tilde{\boldsymbol{z}}_l\mid X \sim \mathcal{N}(\boldsymbol{0},\Sigma_l)$ with
$\Sigma_l=\sum_{r=1}^R K_{XX}^{(r,l)}$ and entries $[K_{XX}^{(r,l)}]_{ij}=k_l^{(r)}(\boldsymbol{x}^{(r)}_i,\boldsymbol{x}^{(r)}_j)$ for $i,j\in\{1,\dots,N\}$.
This additive design lets distinct covariate groups contribute separably to longitudinal dependence.

As in a standard VAE, we use a factorized encoder \quad \( q_\phi(Z\mid Y) = \prod_{i=1}^N q_\phi(\boldsymbol{z}_i \mid \boldsymbol{y}_i) \) and a (Gaussian) decoder \( p_\psi(Y\mid Z) \), and train via
\begin{align}
\label{ELBO_LVAE}
\log p_{\psi,\theta}(Y\mid X)\; & \ge\;
\mathcal{L}_{\phi,\psi,\theta}(Y\mid X)  \\
&=\mathbb{E}_{q_\phi(Z\mid Y)}\!\big[\log p_\psi(Y\mid Z)\big]  \nonumber \\
& \quad - D_{\mathrm{KL}}\!\big(q_\phi(Z\mid Y)\,\|\,p_\theta(Z\mid X)\big), \nonumber 
\end{align}
where \( p_\theta(Z\mid X) \) is the additive GP prior introduced above. We train the model with a low-rank inducing-point approximation, yielding a scalable objective \cite{ramchandran2021longitudinal}. Unlike the i.i.d.\ VAE prior, the LVAE prior explicitly encodes within-subject correlation and temporal dynamics while retaining the same amortized inference machinery.

% The estimated latent representations, either from VAE or LVAE, are further analyzed by the transformer-based sequence aggregation to enhance temporal analysis and improve survival risk predictions. The integration of VAE or LVAE thus plays a central role in enhancing the model's precision and reliability in estimating survival outcomes.
\paragraph{Transformer-based sequence aggregation}
We use an encoder-style \emph{transformer} \cite{vaswani2017attention}, where tokens are the per-visit latent vectors from the VAE/LVAE (optionally concatenated with selected covariates), embedded to a common $d$-dimensional (transformer hidden size) space and augmented with positional/time-gap encodings to preserve order and irregular spacing. Multi-head self-attention captures both short- and long-range dependencies without assuming regular sampling. 
An attention-pooling layer then summarizes the sequence into a single vector, which a small MLP maps to a subject-level risk representation:
\begin{align*}
&H_p = \mathrm{TE}\!\big(\mathrm{Emb}[Z_p,X_p] + \mathrm{PE}\big), \\
&u_p = \mathrm{AttnPool}(H_p),\\
&f_\omega(Z_p,X_p) = \mathrm{MLP}(u_p),
\end{align*}
where $\mathrm{Emb}(\cdot)$ denotes the token embedding layer mapping inputs to $\mathbb{R}^d$, $\mathrm{PE}$ denotes positional/time-gap encodings, $\mathrm{TE}(\cdot)$ is a standard transformer encoder stack, $\mathrm{AttnPool}(\cdot)$ denotes attention pooling, and $\mathrm{MLP}(\cdot)$ is a small multilayer perceptron producing the Cox risk function.

\paragraph{Proportional hazards head}
We model subject–level risk with a Cox formulation \cite{cox1972regression},
\begin{align}
\label{COX_hazard}
h(t \mid Z_p,X_p) \;=\; h_0(t)\,\exp\!\big(f_{\omega}(Z_p,X_p)\big),
\end{align}
where $f_{\omega}$ is the nonlinear predictor produced by the transformer survival head from the VAE/LVAE latent sequence (and selected covariates), and \( h_0(t) \) is the baseline hazard function, representing the hazard for a subject with a baseline level of the covariates. 
% This follows the nonlinear Cox extension of \emph{DeepSurv}, which replaces the linear predictor $\boldsymbol{\beta}^\top \boldsymbol{v}$ with a neural network $f(\boldsymbol{v})$ \cite{katzman2018deepsurv}, where \( \boldsymbol{v} \) represents the predictor variables, and \( \boldsymbol{\beta} \) denotes the coefficients. Here $f_{\omega}$ operates on learned latent trajectories. When $f_{\omega}(Z_p,X_p)=\boldsymbol{\beta}^\top \boldsymbol{v}$, \eqref{COX_hazard} reduces to the classical Cox model.
This corresponds to the standard nonlinear (deep) Cox model popularized by DeepSurv \cite{katzman2018deepsurv}; when $f_{\omega}$ is linear, \eqref{COX_hazard} reduces to the classical Cox model.

\paragraph{Expected Cox partial loss}
We train the survival head by minimizing the expected negative partial log-likelihood under the variational posterior:
\begin{align}\label{eq:expectedpartialloss}
\mathcal{L}_{\omega}(Z,X) =& - \mathbb{E}_{q_{\phi}(Z|Y)}\bigg[  \sum_{p: e_p=1} \bigg( f_{\omega}(Z_p,X_p) \\
& \quad - \log \sum_{j \in \mathcal{R}(t_{p})} \exp(f_{\omega}(Z_j,X_j)) \bigg) \bigg]. \nonumber
\end{align}
where $e_p$ indicates an event and $\mathcal{R}(t_p)$ is the risk set just prior to $t_p$. If $Z$ is deterministic, \eqref{eq:expectedpartialloss} collapses to the standard Cox partial loss.

\paragraph{Loss Function}
\label{subsec:loss_function}

The loss function of the SeqRisk is designed to simultaneously optimize the hazard regression and ELBO of VAE objective. 
It integrates a risk regularization parameter to balance the survival analysis objectives with the generative modeling capabilities of the VAEs. 
The composite loss function is defined as
\begin{align*}
\mathcal{L}_{\phi, \psi, \theta, \omega}(Y |  X) &= \alpha \mathcal{L}_{\omega}(Z,X)  - (1-\alpha) \mathcal{L}^{\text{elbo}}(Y |  X),
\end{align*}
where $\mathcal{L}_{\omega}(Z,X)$ denotes the expected negative partial log-likelihood of hazard regression (which we define in more details in Equation \eqref{eq:expectedpartialloss} above) on the latent representations $Z$ and covariates $X$ %
and $\mathcal{L}^{\text{elbo}}(Y |  X)$ denotes the ELBO, which aids in the effective generative modeling of the data. The ELBO term corresponds to $\mathcal{L}_{\phi, \psi}(Y)$ for the standard VAE model as given in Equation \eqref{ELBO_VAE} or $\mathcal{L}_{\phi, \psi,\theta}(Y |  X)$ for LVAE as given in Equation \eqref{ELBO_LVAE}. 
The risk regularization parameter $\alpha$, chosen by cross-validation, balances survival prediction accuracy with latent representation quality. 
This loss thus captures SeqRisk’s dual aim: accurate time-to-event prediction and effective modeling of longitudinal structure.

\subsection{Evaluation Using Time-Independent Concordance Index}
\label{subsec:time_independent_cindex}

We evaluate SeqRisk using the time-independent concordance index (C-index), a standard metric for assessing survival model performance \cite{harrell1982evaluating}. It measures how well the model ranks patients by risk, offering a simple summary of predictive accuracy widely accepted in clinical practice.

Let $\hat R_p := f_{\omega}(Z_p,X_p)$ denote the predicted risk score for subject $p$, the C-index is defined as:
\begin{align*}
\text{C-index} =
\frac{\sum_{p}\sum_{q} \mathds{1}(t_p < t_q,\, e_p=1)\,\mathds{1}(\hat R_p > \hat R_q)}
{\sum_{p}\sum_{q} \mathds{1}(t_p < t_q,\, e_p=1)}.
\end{align*}
where \( t_p \), \( t_q \) are observed survival times, \( e_p \) indicates event occurrence, and \( \mathds{1} \) is the indicator function. While it does not capture time-varying risk, the C-index remains practical for summarizing model performance in real-world clinical settings.

\section{Experiments}
\label{sec:theorems}
We evaluate the SeqRisk model using the concordance index (C-index) to compare its predictive performance against established methods with three datasets. Integrated Brier score (IBS) \cite{graf1999assessment} was evaluated only on PhysioNet due to resource constraints for CHD and limited interpretability for synthetic Survival MNIST data.

\subsection{Datasets}
% \label{subsec:datasets}

\subsubsection{Survival MNIST Dataset}
\label{subsubsec:survival-mnist-synthetic-dataset}
The Survival MNIST synthetic dataset simulates disease progression using MNIST digit images ($36 \times 36$ pixels), where each individual is represented by a different MNIST digit number three, and the progression is modeled as a gradual rotation of the images, with $0^\circ$ representing a healthy state and $180^\circ$ representing the disease endpoint. The dataset generation process includes several steps to emulate real-world healthcare data characteristics: adding Gaussian noise, masking 70\%, 80\%, 90\%, 95\% and 99\% of the pixels to reflect data sparsity, and randomly designating each image as either having experienced the event with 60\% of probability or being censored otherwise. 
The number of observation points for each digit is randomly chosen between 5 and 20 to simulate irregular sampling times. For this dataset, observed images are the measurements and observation times as well as ids of the subjects are the covariates.

This synthetic dataset provides a controlled environment to evaluate the model's ability to handle the complexities inherent in longitudinal survival data, including noise, sparsity, irregular observation intervals, and censoring. An example observed sequence with different amounts of missingness is provided in Appendix \ref{subsubsec:appendix-survival-mnist-synthetic-dataset}.
%Figure \ref{fig:example_mnist_v2}.

\subsubsection{PhysioNet Challenge 2012 dataset}
\label{subsubsec:real-dataset-physionet}

We evaluate our model on the PhysioNet Challenge 2012 dataset, which contains multivariate time-series data from Intensive Care Unit (ICU) stays \cite{silva2012predicting}. Each patient's clinical measurements—including vital signs, laboratory results, and demographic variables—are recorded over the first 48 hours of ICU admission at irregular intervals. The prediction target is in-hospital mortality after the initial 48-hour observation window.

The dataset comprises a total of 11,976 patients, partitioned into 6,385 for training, 1,597 for validation, and 3,994 for testing. Patients have a median of 72 observations, covering 37 distinct clinical measurements, in addition to demographic attributes such as ICU type, gender, and age. 
%Patients with insufficient measurements during the 48-hour window are excluded.

Missing values are explicitly retained for neural models using masking indicators, while for non-neural baselines, we apply mean imputation, which we found to yield better performance. The final cohort is characterized by high sparsity, with more than 84\% missingness across features, reflecting real-world variability in ICU data collection. Despite this, the early physiological patterns captured during the first 48 hours provide meaningful signals for risk prediction.

\subsubsection{Coronary heart disease dataset}
\label{subsubsec:real-dataset-hus}

We use twelve-year follow-up data of patients with coronary heart disease (CHD) from HUS Helsinki University Hospital. 
From the full health records, only laboratory measurements and demographic variables were retained for mortality prediction. 
Preprocessing included (i) monthly aggregation using medians, (ii) exclusion of patients with fewer than ten time sequences, and (iii) removal of rare lab tests, requiring at least one occurrence in training. 
This reduced the dataset from 987 to 685 lab tests, 5101 to 4058 patients, and 409{,}116 to 128{,}018 observations. 
Patient sequences have a median length of 27 (min.~10, max.~139). 
During follow-up, 18.4\% of patients experienced the event and the rest were right-censored. 

After preprocessing, overall missingness was 98.34\% (the least-missing test had 23\% missingness). 
To assess robustness, we further subsampled lab tests with less than 70\%, 80\%, and 90\% missingness, yielding datasets with 98.91\%, 99.08\%, and 99.31\% overall missingness. 
This simulates the irregular availability of lab tests in real-world EHRs and provides increasingly challenging sparsity scenarios. 
Details of preprocessing are provided in Appendix \ref{subsubsec:appendix-real-dataset-hus}.

\subsection{Baseline Models}
 \label{subsec:baseline_models}
To evaluate SeqRisk, we benchmarked against both classical and modern survival models. As classical baselines, we included the \textbf{Cox proportional hazards model (Cox)} \cite{cox1972regression} and \textbf{Random Survival Forests (RSF)} \cite{ishwaran2008random}, widely used for their robustness to censoring and nonlinearity. For recent deep learning methods on longitudinal data, we compared to \textbf{Dynamic DeepHit (DDH)} \cite{lee2019dynamic}, which uses recurrent networks with temporal attention to capture time-dependent patterns, and \textbf{Recurrent Deep Survival Machines (RDSM)} \cite{nagpal2021recurrent}, which models time-varying covariates and survival times as mixtures of parametric distributions. These baselines were evaluated at dataset-specific time horizons chosen for high event density, ensuring fair comparison.  

In addition to these established models, we consider four variants of our SeqRisk architecture as internal baselines. The first variant utilizes the VAE for latent space representation followed by a multilayer perceptron (MLP) for risk regression using only the last time point for each individual (\textbf{SeqRisk: VAE+MLP}). This variant helps isolate the impact of replacing the transformer-based module with a more traditional neural network in the survival prediction task. Another internal baseline is a variant that excludes the VAE component (\textbf{SeqRisk: Transformer only}, which directly applies a transformer-based module to the observed covariates and measurements, encoding the longitudinal data without any intermediate latent representation. This baseline is crucial for evaluating the contribution of the VAE/LVAE component to overall model performance. The third and fourth variants correspond to our full proposed methods as described in Section \ref{sec:seqrisk_framework}: \textbf{SeqRisk: VAE+Transformer} and \textbf{SeqRisk: LVAE+Transformer}, respectively.
SeqRisk was implemented in PyTorch (2.1.0); the code and Conda environment specification are available at \href{https://github.com/MineOgre/SeqRisk}{github.com/MineOgre/SeqRisk}.

% Finally, we note that certain other advanced models from related work were considered but ultimately not included in our baseline comparison. For example, the neural ODE-based survival model \textbf{SurvLatent ODE} \cite{moon2022survlatent} and a recent variational deep survival clustering approach \cite{manduchi2021deep} have been proposed for longitudinal survival data. However, the extremely high dimensionality and sparsity of our datasets—particularly the CHD cohort with hundreds of covariates and over 98\% missingness—made these approaches impractical to implement as baselines. 

We did not include some advanced longitudinal survival baselines (e.g., SurvLatent ODE \cite{moon2022survlatent} and variational survival clustering \cite{manduchi2021deep}) because the extreme dimensionality and sparsity of our data (CHD: hundreds of covariates, $>98\%$ missingness) made them impractical to implement.

%Such models would face prohibitive computational requirements or convergence difficulties on data of this scale.
% , so we focus our evaluation on the more tractable methods.

\subsection{Experiment Setup}

For neural network-based models, we segregated the datasets into training, validation, and testing sets to facilitate rigorous model training and unbiased evaluation, retaining the original PhysioNet dataset splits provided by the PhysioNet Challenge 2012 for comparability. 
% For Cox model, the validation data were incorporated into the training set. 
% For Cox, we merged the training and validation splits and trained the model on the combined data. 
% For RSF, the combined training–validation set was used for hyperparameter selection via grid search with cross-validation, after which the model was retrained on the same data with the selected parameters.

For the MNIST dataset, we utilized three random splits to ensure robustness and reliability in our model evaluations. For the CHD dataset, we implemented a 5-fold cross-validation to ensure that each subject is included in the test set exactly once. The final results for both datasets are reported as the mean and standard deviation of the C-index across all splits. For the PhysioNet dataset, we report the test results of the experiment that yield the best validation performance.

In our experiments, missing data for Cox and RSF models were imputed using two distinct imputation techniques: mean and K-nearest neighbors (KNN) imputations. Given these models' static nature, we utilized only the last observation time point. Additionally, for RSF, we conducted a grid search to fine-tune various parameters, ensuring optimal model configuration. We report the best-performing approach in the final evaluations for each model. For the neural network-based models, we experimented with various configurations and selected the best validation scores to determine the final test results.

\paragraph{Implementation}
For the MNIST dataset, convolutional encoder and decoder networks were employed, optimized specifically for image data. 
The SeqRisk: LVAE+Transformer model for this dataset incorporated Gaussian process (GP) configurations accounting for observation time, patient ID, and their interactions. For the CHD dataset, GP configurations in the SeqRisk: LVAE+Transformer model included patient ID, observation time, age, and interactions between observation time and various factors such as gender, treatment plan, arrhythmia, and smoking status. For the PhysioNet dataset, the GP configurations considered patient ID and observation time. 
Detailed model configurations are available in Appendix \ref{subsec:Seqrisk_impl}.

\subsection{Results}

\begin{figure}[htbp]
  \centering
  \vspace{-15pt}
  \includegraphics[width=\linewidth]{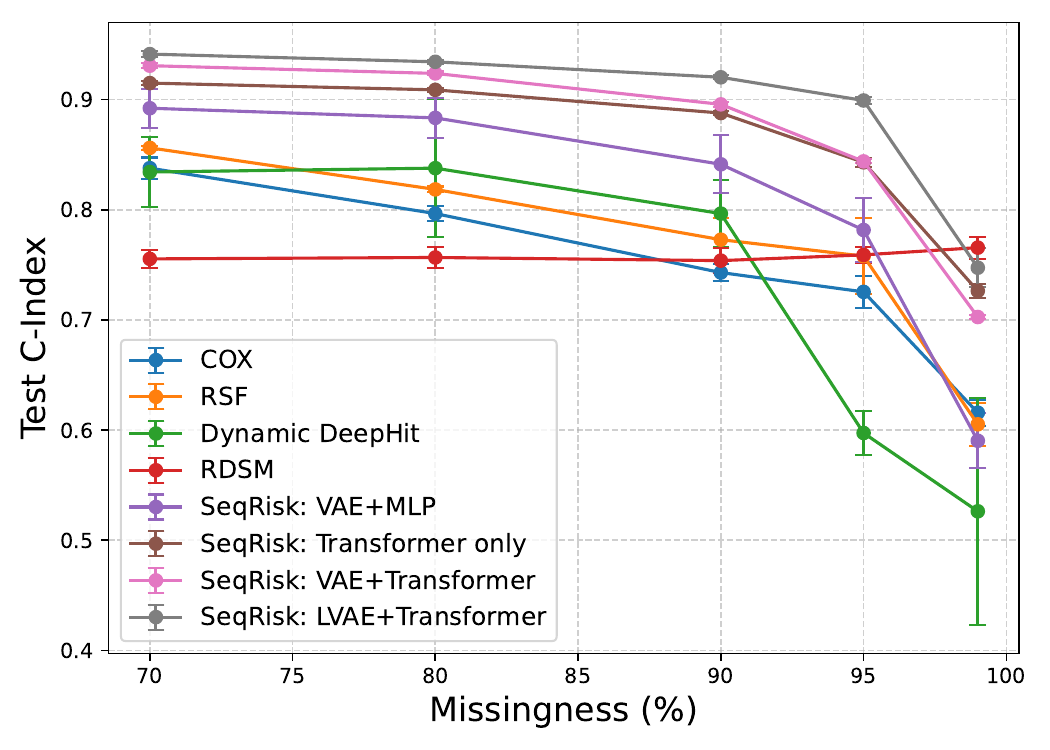}
  \caption{Test C-index scores of Survival MNIST with varying amounts of missingness}
  \label{fig:mnist_plot}
\end{figure}

Model performances on the Survival MNIST dataset are reported in Figure \ref{fig:mnist_plot}. 
SeqRisk variants with transformer aggregation achieve the highest C-indices at low missingness, with SeqRisk: LVAE+Transformer, substantially outperforming Cox and RSF over the same range. 
Even at 99\% missingness, SeqRisk: LVAE+Transformer and VAE+Transformer remain competitive, whereas SeqRisk: VAE+MLP collapses and DDH falls sharply at 99\%. 
RDSM is distinctive in showing an almost flat curve across all levels, reflecting its explicit modeling of the event--time distribution through a parametric mixture \cite{nagpal2021recurrent}. 
Given that Survival MNIST event times are generated from a simple monotone rotation process, RDSM can capture this distribution directly, explaining its robustness to sparsity but also its limited accuracy relative to SeqRisk.

\begin{table}[t]
  \centering
  \caption{Test C-Index for PhysioNet 2012}
  \label{tab:physionet_cindex}
  \begin{tabular}{ll}
    \toprule
    Method                           & Test C-Index \\
    \midrule
    Cox                              & 0.660        \\
    RSF                              & 0.679        \\
Dynamic DeepHit       (DDH)           & 0.779        \\
    RDSM                             & 0.749       \\
 \hline
    SeqRisk: Transformer only         & 0.770        \\
    SeqRisk: VAE+MLP                 & 0.686        \\ \hline
    SeqRisk: VAE+Transformer         & \textbf{0.789}        \\
    SeqRisk: LVAE+Transformer        & 0.765        \\

    \bottomrule
  \end{tabular}
\end{table}

\begin{table*}[h]
\centering
\caption{Test C-index scores for Coronary Heart Disease (CHD) Dataset with varying amounts of missingness.}
\label{tab:conc_index}
\small
\begin{tabular}{lcccc}
\hline
\hline
Model & \multicolumn{4}{c}{Missingness for the least sparse lab test (Overall Missingness) \%} \\
\cline{2-5}
      & 23 (98.34) & 70 (98.91) & 80 (99.08) & 90 (99.31) \\
\hline
\hline
Cox                                   & $0.688 \pm 0.025$ & $0.695 \pm 0.010$ & $0.678 \pm 0.023$ & $0.690 \pm 0.016$ \\
RSF                                   & $\mathbf{0.867 \pm 0.017}$ & $0.854 \pm 0.018$ & $0.847 \pm 0.021$ & $0.839 \pm 0.020$ \\
Dynamic DeepHit        (DDH)                & $0.782 \pm 0.023$ & $0.769 \pm 0.024$ & $0.764 \pm 0.018$ & $0.759 \pm 0.018$ \\
RDSM                                   & $0.756 \pm 0.017$ & $0.713 \pm 0.025$ & $ 0.707\pm 0.030$ & $ 0.715\pm 0.041$ \\
 \hline
SeqRisk: Transformer only& $0.846 \pm 0.018$ & $0.845 \pm 0.021$ & $0.847 \pm 0.021$ & $0.841 \pm 0.017$ \\
SeqRisk: VAE+MLP             & $0.841 \pm 0.025$ & $0.835 \pm 0.016$ & $0.799 \pm 0.017$ & $0.806 \pm 0.011$ \\ \hline
SeqRisk: VAE+Transformer         & $\mathbf{0.871 \pm 0.015}$ & $\mathbf{0.874 \pm 0.014}$ & $\mathbf{0.876 \pm 0.015}$ & $\mathbf{0.872 \pm 0.016}$ \\
SeqRisk: LVAE+Transformer   & $0.853 \pm 0.014$ & $0.847 \pm 0.012$ & $0.849 \pm 0.012$ & $0.848 \pm 0.012$ \\
\hline
\end{tabular}
\end{table*}

\begin{table}[ht]
\centering
\caption{Integrated Brier Score (IBS) on the PhysioNet dataset. Lower values indicate better calibration.}
\label{tab:ibs_physionet}
\begin{tabular}{lr}
\toprule
\textbf{Model} & \textbf{IBS Score} \\
\midrule
Cox  & 0.193 \\
RSF & 0.211 \\
Dynamic DeepHit (DDH)$^{\dagger}$ & \textbf{0.081} \\
RDSM & 0.182 \\ \hline
SeqRisk: Transformer only & 0.144 \\
SeqRisk: VAE + MLP & 0.187 \\ \hline
SeqRisk: VAE + Transformer & 0.142 \\
SeqRisk: LVAE + Transformer$^{\ddagger}$ & \underline{0.135} \\
\midrule
% \multicolumn{2}{p{0.9\linewidth}}{\scriptsize \textit{Note:} IBS was evaluated only on PhysioNet due to resource constraints for CHD and limited interpretability for synthetic Survival MNIST data.} \\
\multicolumn{2}{p{0.9\linewidth}}{\scriptsize $^{\dagger}$ Best overall calibration due to DDH’s full event-time modeling.\par  $^{\ddagger}$ Proposed SeqRisk model; competitive calibration and robust to irregular, noisy longitudinal data.} \\
\bottomrule
\end{tabular}
\end{table}

Performance of different models for the PhysioNet dataset are shown in Table \ref{tab:physionet_cindex}. Our results show that SeqRisk: VAE+Transformer achieves the best overall performance, with DDH also performing competitively. Consistent with observations on other datasets, SeqRisk models incorporating a transformer module demonstrate strong predictive capability, reinforcing the effectiveness of transformer-based architectures. Given the clinical urgency of accurate ICU mortality predictions, the enhanced predictive power of SeqRisk could translate into more timely interventions and improved patient outcomes. The configurations of the models in the table with best validation performances are given in Appendix \ref{subsec:Seqrisk_impl}.

On the CHD dataset, SeqRisk emerges as a strong performer across varying levels of missing data. 
% While RSF attains slightly higher performance on the dataset with the least missingness (Table \ref{tab:conc_index}), 
SeqRisk demonstrates robust generalization as sparsity increases ( Table \ref{tab:conc_index}). 
In particular, DDH, SeqRisk: VAE+MLP, and RSF exhibit a regular decline as data becomes sparser, whereas SeqRisk variants maintain more stable performance, underscoring their suitability for longitudinal patient monitoring where data completeness varies substantially. 
Compared with RSF as the primary non–SeqRisk comparator, our best variant (SeqRisk: VAE+Transformer) achieves consistently higher C-index under sparse settings. % (Appendix, Table~\ref{tab:chd_rsf_vs_seq}). 
At 10\%, 20\%, and 30\% missingness these gains are statistically significant, with 
$\Delta C{=}0.033$ (95\% CI $[0.011,\,0.055]$), 
$\Delta C{=}0.029$ ($[0.010,\,0.047]$), and 
$\Delta C{=}0.021$ ($[0.006,\,0.036]$), respectively (paired $t$-tests, $p<0.05$). 
On the full dataset, the difference is small and not significant ($\Delta C{=}0.004$, $[{-}0.005,\,0.014]$). 
Overall, these results highlight SeqRisk’s ability to degrade more gracefully under extreme sparsity, providing a clear advantage in the high-missingness regimes characteristic of longitudinal EHR data.

To complement the C-index, we evaluate the \textit{Integrated Brier Score (IBS)} on the PhysioNet dataset (Table~\ref{tab:ibs_physionet}), which measures the calibration of predicted survival probabilities. Dynamic DeepHit (DDH) achieves the lowest IBS due to its explicit modeling of the full event-time distribution, yielding well-calibrated survival estimates. 
% SeqRisk: LVAE+Transformer, while optimized primarily for discrimination via a Cox-based partial likelihood objective, outperforms classical baselines and achieves competitive calibration among deep learning models. 
Although optimized for discrimination, SeqRisk: LVAE+Transformer outperforms classical baselines and shows competitive calibration among deep models.

% We report IBS only for PhysioNet due to resource constraints for the CHD cohort and the limited interpretability of probabilistic calibration on synthetic datasets like Survival MNIST.

\begin{figure}[htbp]
  \centering
  \vspace{-5pt}
  \includegraphics[width=.9\linewidth]{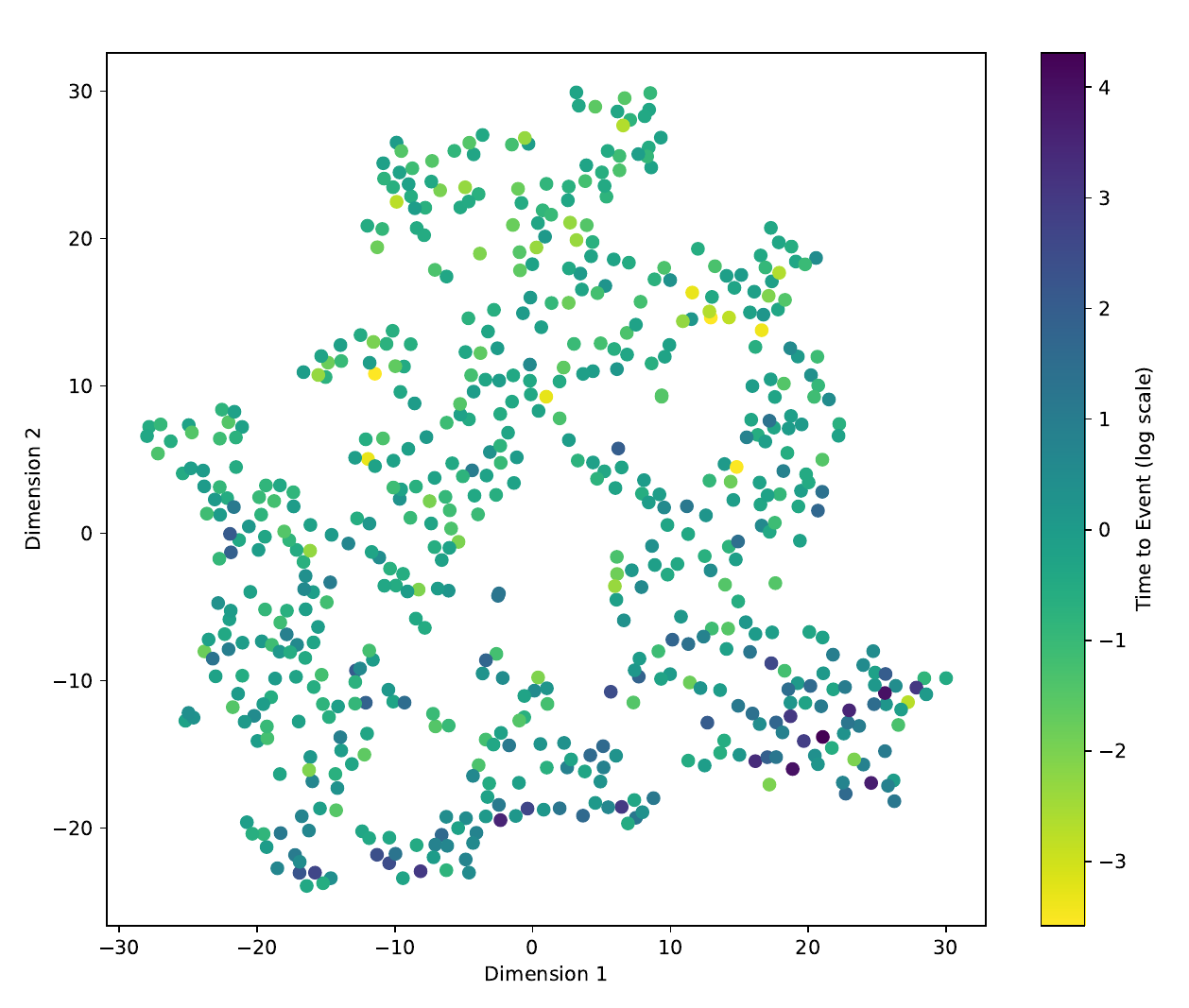}
  \caption{UMAP-based 2D visualization of latent embeddings and selected covariates; colors represent time-to-event (log scale), providing global but not feature-specific interpretability.}
  \label{fig:scatter_plot}
\end{figure}

The scatter plot in Figure \ref{fig:scatter_plot} 
presents a UMAP-based \cite{mcinnes2018umap} two-dimensional visualization of the VAE latent representation combined with selected covariates from the CHD dataset, colored according to time-to-event in log scale. A clear structure of latent space is visible where embeddings associated with lower-risk patients tend to cluster towards the lower part of the plot. Although explicit clinical factors influencing risk aren't directly interpretable from this plot, the clear separation indicates SeqRisk effectively encodes clinically meaningful temporal patterns. Further interpretability efforts might include deeper analysis of clinical variables contributing to these embeddings.

\section{Discussion}
Our experiments demonstrate that SeqRisk effectively enhances survival prediction by combining latent trajectories from a VAE/LVAE with a transformer-based sequence aggregation. A key strength of SeqRisk is its robustness in handling highly sparse and irregularly sampled longitudinal data, which are prevalent in clinical practice due to missed visits, varied measurement schedules.

SeqRisk: LVAE+Transformer is the most robust variant on Survival MNIST under high missingness and achieves the second-best calibration on PhysioNet, surpassed only by Dynamic DeepHit (DDH), which benefits from full event-time modeling. However, LVAE+Transformer does not consistently improve discrimination (C-index) over VAE+Transformer on the real-world datasets. Meanwhile, SeqRisk: VAE+Transformer performs competitively across settings—closely matching RSF on the CHD dataset and DDH on PhysioNet—highlighting the framework’s adaptability to heterogeneous clinical contexts. Overall, SeqRisk can improve predictive accuracy and risk calibration over classical survival methods and simpler deep-learning baselines, with potential impact on clinical decision-making in chronic disease management, ICU monitoring, and personalized healthcare planning.

Visualization of latent patient embeddings reveals risk groups, although our current interpretability is limited to this global risk perspective rather than explicit identification of influential clinical factors. Future work incorporating explicit interpretability methods, such as transformer attention weight analysis, could further improve clinical usability.

We selected the C-index due to its widespread adoption in survival analysis and its intuitive interpretation as a measure of how well the model ranks patients by risk. Accurate ranking of patient risks is particularly relevant for identifying high-risk individuals and guiding early clinical interventions. While our current SeqRisk model focuses on single-event outcomes, extending it to address competing risks or multi-event scenarios through multi-task or competing-risk frameworks would enhance its applicability in clinical contexts. Additionally, integrating heterogeneous likelihood methods tailored to diverse clinical data types or compositional data \cite{ogretir2022hetero, ogretir2023longitudinal} presents another promising direction to improve model generalizability.

Overall, SeqRisk demonstrates potential to advance survival analysis in healthcare by effectively addressing common real-world data challenges and providing robust and accurate risk predictions.

% geri eklenece

% \acks{Acknowledgments go here \emph{but should only appear in the
% camera-ready version of the paper if it is accepted}.
% % Acknowledgments do not count toward the paper page limit.}

%%%%%%%%%%%%%%%%%%%%%%%%%%%%%%%%%%%%%%%%%%%%%%%%%%%%%%%%%%%%%%%%%%%%%%%%%%%%%%%%
\section*{APPENDIX}

\section{Detailed Dataset Information}
\label{sec:appendix-datasets}

\subsection{Survival MNIST Synthetic Dataset}
\label{subsubsec:appendix-survival-mnist-synthetic-dataset}

The Survival MNIST Synthetic Dataset simulates disease progression using MNIST digits, where disease severity is modeled as a gradual rotation from $0^\circ$ (healthy) to $180^\circ$ (terminal).

\subsubsection{Data Generation Process}
The dataset reflects key characteristics of clinical data—noise, sparsity, and irregular sampling. Each digit is rotated to represent progression. To emulate noise, images are randomly shifted along the diagonal and perturbed with Gaussian noise $\mathcal{N}(0,30)$. Pixel values are masked at rates of 70–99\% to induce sparsity. Each subject is randomly assigned as an event (0.6) or censored (0.4). The number of observation points per subject is uniformly sampled between 5 and 20; the last observation lies in the second half of the progression timeline, with times uniformly distributed between start and end. This design creates a controlled yet realistic setting with noise, missingness, and censoring.

\subsubsection{Experiment Setup}
\label{subsubsec:experiment-setup}
The dataset is split into 60\% training, 20\% validation, and 20\% test sets. Three distinct data splits assess robustness, and each experiment is repeated three times with different seeds. For SeqRisk (VAE and LVAE+Transformer), the VAE/LVAE is trained without outcome supervision to learn a latent representation of covariate trajectories, while hazard regression is trained only on the training set using survival outcomes.

% For SeqRisk (VAE and LVAE+Transformer), the full dataset trains the VAE to capture the overall distribution, while only the training survival data guide hazard regression to ensure proper generalization. This setup provides a compact yet rigorous evaluation of SeqRisk under realistic survival analysis challenges.

\subsection{Coronary heart disease (CHD) dataset}
\label{subsubsec:appendix-real-dataset-hus}

The real-world dataset employed in this study originates from a comprehensive longitudinal study conducted over 12 years, focusing on patients with coronary heart disease.
%at Helsinki University Hospital. 
The study, detailed in the International Journal of Epidemiology \cite{vaara2012cohort}, 
provides an extensive dataset encompassing various aspects of patient health and disease progression.

\subsubsection{Dataset Description and Selection}
We used the laboratory measurements segment of the dataset, which includes a wide range of biomarkers relevant to coronary health and disease. 
These measurements provide critical insights into the biological processes and risk factors associated with coronary heart disease, making them invaluable for our analysis.

\subsubsection{Data Preprocessing}
To ensure data quality and longitudinal utility, laboratory measurements were aggregated monthly using median values to reduce timestamp granularity and increase data density. We excluded patients with fewer than 10 observations to retain individuals with sufficient longitudinal information. Finally, within each 5-fold split, we retained only lab tests that appear in the corresponding training fold, avoiding failures caused by features that are entirely missing at training time.

\subsubsection{Experiment Setup}
\label{subsec:experiment-results-coronary-data}

CHD dataset was used to validate the SeqRisk framework in a real-world clinical setting. The data were partitioned into 60\% training, 20\% validation, and 20\% testing, consistently applied across 5-fold splits to ensure each subject appeared in the test  once. Each split was repeated three times to assess stability, except for the deterministic Cox model.
Both latent variable models (SeqRisk: VAE and SeqRisk: LVAE) were trained on the full dataset to maximize latent representation quality. The survival head was trained only on training subjects.

This real-world dataset enables validation of the SeqRisk’s utility in clinical applications. Modeling lab trajectories over 12 years demonstrates ability to handle high-dimensional, 
sparse, and irregular data while capturing disease progression.

\section{Implementation Details}
\label{subsec:Seqrisk_impl}

% The model configurations for the Survival MNIST dataset and CHD dataset are summarized in Table \ref{tab:model_configs}. The detailed convolutional NN configuration for Survival MNIST dataset is given in Table \ref{tab:neural_network_architecture}. The experiments run with 4, 8, 16, 32 and 64 dimension in latent representation for Survival MNIST dataset, and with 8, 16 and 32 for CHD dataset. 

The model configurations for the MNIST and CHD datasets differ in several aspects. For MNIST, the transformer uses 2 layers with 2 heads, a feed-forward width of 2, and a convolutional encoder/decoder network. For CHD, the transformer has 1 layer with 4 heads, a feed-forward width of 4, and encoder/decoder layers of sizes [200, 50] and [50, 200]. Both models include a single-layer MLP with 50 dimensions after attention. Gaussian process covariates include time and patient ID for MNIST, and for CHD, ID, time, age, and interaction terms such as time × gender, time × treatment, time × arrhythmia, and time × smoking.

The Convolutional Neural Network used in Survival MNIST takes \(36\times36\) inputs. The inference network consists of convolutional layers with 32 filters, \(3\times3\) kernels, stride 1, and max pooling with \(2\times2\) kernels and stride 2. It includes two feedforward layers of widths 300 and 30, uses ReLU activations, and maps to a latent space of dimension \(L\). The generative network takes \(L\)-dimensional inputs, has two transposed convolutional layers with 16 filters, \(4\times4\) kernels, stride 2, and two feedforward layers of widths 30 and 300, also with ReLU activations.

The experiments run with 4, 8, 16, 32 and 64 dimension in latent representation for Survival MNIST dataset, and with 8, 16 and 32 for CHD dataset. 
The choice of the risk regularization parameter, \(\alpha\), is critical in balancing the emphasis between the survival prediction accuracy and the robustness of the latent space representation. As a hyperparameter, \(\alpha\) is selected through a systematic hyperparameter tuning process that seeks to optimize model performance with cross validation. 

\section*{Acknowledgment}
This study was supported by VTR Research Funds from the HUS, Helsinki University Hospital, Helsinki, Finland, and the Research Council of Finland (decision \#359135).

\end{document}